\documentclass[lettersize,journal]{IEEEtran}
\usepackage{amsmath,amsfonts}
\usepackage{algorithm}
\usepackage{array}
\usepackage[caption=false,font=normalsize,labelfont=sf,textfont=sf]{subfig}
\usepackage{textcomp}
\usepackage{stfloats}
\usepackage{url}
\usepackage{verbatim}
\usepackage{graphicx}
\def\BibTeX{{\rm B\kern-.05em{\sc i\kern-.025em b}\kern-.08em
    T\kern-.1667em\lower.7ex\hbox{E}\kern-.125emX}}
\usepackage{balance}
\usepackage{multirow}
\usepackage{textcomp}
\usepackage[table]{xcolor}
\usepackage{booktabs}
\usepackage{makecell}
\usepackage{pifont} 

\begin{document}
\title{QuadINR: Hardware-Efficient Implicit Neural Representations Through Quadratic Activation}
\author{Wenyong Zhou\textsuperscript{\dag}, Boyu Li\textsuperscript{\dag}, Jiachen Ren, Taiqiang Wu, Zhilin Ai, Zhengwu Liu*, and Ngai Wong*
\thanks{This research was partially conducted by ACCESS – AI Chip Center for Emerging Smart Systems, supported by the InnoHK initiative of the Innovation and Technology Commission of the Hong Kong Special Administrative Region Government, and partially supported by the Theme-based Research Scheme (TRS) project T45-701/22-R, the General Research Fund (GRF) Project 17203224 of the Research Grants Council (RGC), Hong Kong SAR, and the National Natural Science Foundation of China Project 62404187.

W. Zhou, B. Li, J. Ren, T. Wu, Z. Ai, Z. Liu and N. Wong are with the Department of Electrical and Electronic Engineering, The University of Hong Kong. \textsuperscript{\dag}: Equal contributions. *: Corresponding authors: Zhengwu Liu and Ngai Wong \{zwliu, nwong@eee.hku.hk\}.}}

\markboth{IEEE Transactions on Circuits AND Systems—II: Express Briefs,~Vol.~XX, No.~X, XXX~2025}%
{How to Use the IEEEtran \LaTeX \ Templates}

\maketitle

\begin{abstract}
Implicit Neural Representations (INRs) encode discrete signals continuously while addressing spectral bias through activation functions (AFs). Previous approaches mitigate this bias by employing complex AFs, which often incur significant hardware overhead. To tackle this challenge, we introduce QuadINR, a hardware-efficient INR that utilizes piecewise quadratic AFs to achieve superior performance with dramatic reductions in hardware consumption. The quadratic functions encompass rich harmonic content in their Fourier series, delivering enhanced expressivity for high-frequency signals, as verified through Neural Tangent Kernel (NTK) analysis. We develop a unified $N$-stage pipeline framework that facilitates efficient hardware implementation of various AFs in INRs. We demonstrate FPGA implementations on the VCU128 platform and an ASIC implementation in a 28nm process. Experiments across images and videos show that QuadINR achieves up to 2.06dB PSNR improvement over prior work, with an area of only 1914$\mu$m$^2$ and a dynamic power of 6.14mW, reducing resource and power consumption by up to 97\% and improving latency by up to 93\% vs existing baselines.
\end{abstract}


\section{Introduction}
\IEEEPARstart{I}{mplicit} Neural Representations (INRs) have emerged as a powerful framework for representing discrete signals, images, audio, and 3D shapes, in a continuous and resolution-independent manner~\cite{deepsdf, chen2021nerv}. This inherent flexibility makes INRs particularly valuable for super-resolution, signal compression and 3D reconstruction applications~\cite{occupancynetworks, mildenhall2020nerf}. At their core, INRs encode signals into Multi-Layer Perceptron (MLP) parameters and leverage non-linear activation functions (AFs) to capture high-frequency details and intricate signal variations with remarkable precision~\cite{siren, wire, gauss, finer}.

\begin{figure}[!t]
\centering
\includegraphics[scale=0.21]{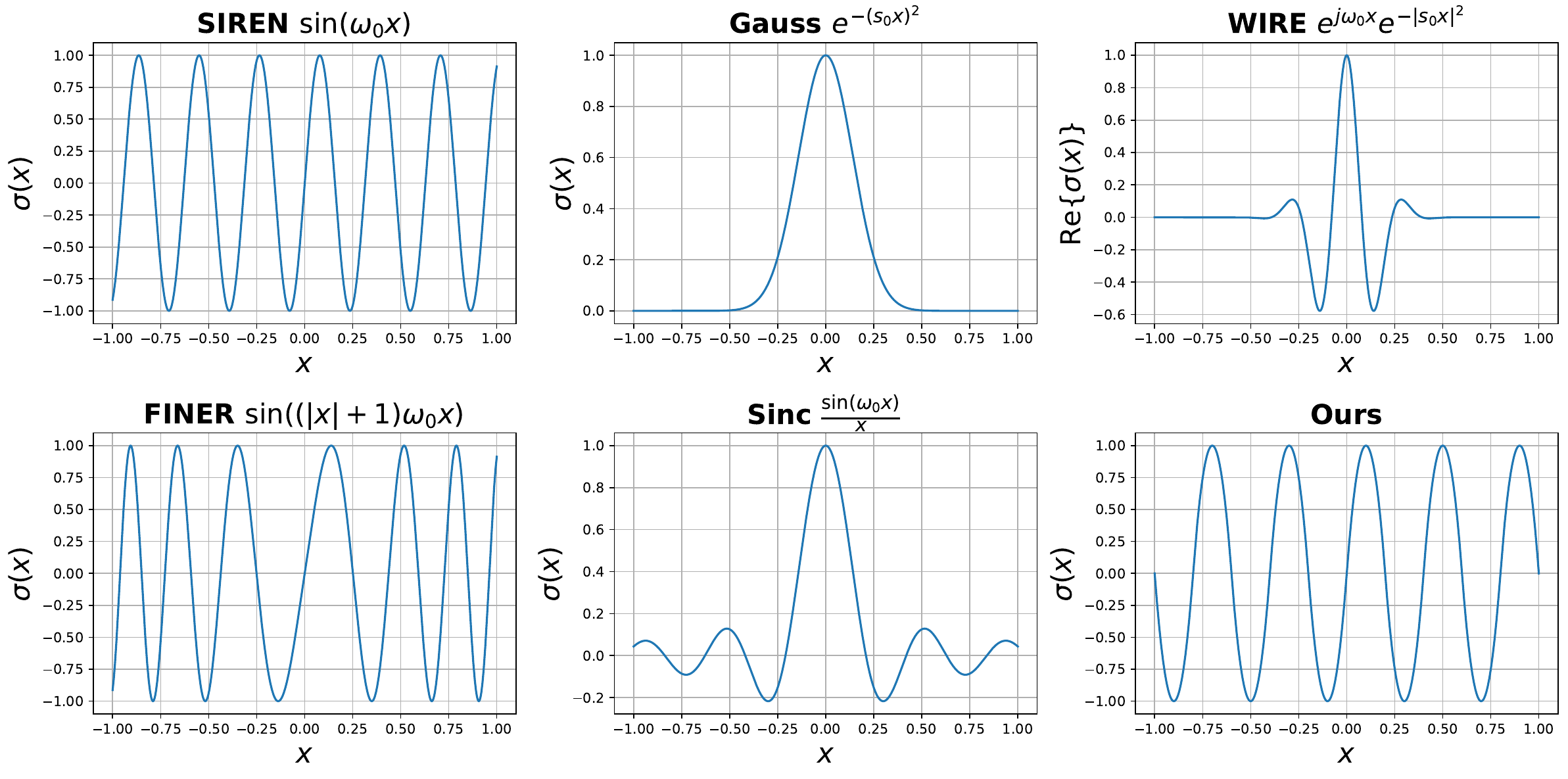}
\caption{Comparison of various non-linear AFs~\cite{siren, wire, sinc, finer, gauss}, including sinusoidal, Gaussian, wavelet, variable periodic, sinc, and our proposed piecewise quadratic functions.}
\label{fig:function}
\end{figure}

A fundamental challenge in INR training is mitigating spectral bias, where neural networks naturally favor learning lower frequency components~\cite{siren, wire}. To address this issue, advanced activation functions (AFs) have been developed to capture a broader range of frequency components, as shown in Fig.~\ref{fig:function}. For example, sinusoidal activations, as used in architectures like SIREN and FINER~\cite{siren, finer}, excel at modeling high-frequency details by leveraging trigonometric functions. Similarly, Gaussian-based functions and wavelet-based functions improve expressivity by incorporating localized frequency responses~\cite{wire, pin}. Recent advances in modified ReLU functions have demonstrated significant potential in enhancing the expressivity of INRs~\cite{bw-relu}, showing promising results that pave the way for their application to increasingly complex and large-scale datasets in future research. 

However, most existing AFs in INRs present notable challenges for hardware implementation. Sinusoidal activations require either Look-Up Tables (LUTs)~\cite{LUT1, LUT2}, iterative approximation methods such as the Coordinate Rotation Digital Computer (CORDIC) algorithm~\cite{CORDIC1, CORDIC2}, or Taylor series expansions~\cite{taylor_1, taylor_2}, each introducing trade-offs between accuracy, latency, and resource utilization~\cite{taylor_2, LUT3, CORDIC3}. Gaussian-based functions rely on exponential computations that typically demand specialized circuits with corresponding hardware requirements. Wavelet-based functions combine elements of sine and Gaussian computations, which can increase implementation complexity in hardware contexts. Similarly, sinc functions~\cite{sinc} require division operations that add another dimension of computational considerations.

To address these limitations, we propose QuadINR, a hardware-efficient INR framework that utilizes piecewise quadratic activations. Theoretical analysis grounded in Fourier transformation and Neural Tangent Kernel (NTK) theory demonstrates that our quadratic function efficiently captures high-frequency components in reconstruction signals while naturally emphasizing low-to-mid-frequency signals during network evolution~\cite{ntk1}. Furthermore, we develop a unified, $N$-stage pipeline framework that facilitates efficient and flexible hardware implementation of various AFs in INRs. We demonstrate FPGA implementations of INRs on the VCU128 platform and an ASIC implementation in a 28\,nm process. In summary, our contributions are three-fold:
\begin{itemize} 
    \item We propose QuadINR, a hardware-efficient INR utilizing a piecewise quadratic activation function. Fourier analysis demonstrates its capacity to capture a broad spectrum of frequency components for fine detail reconstruction, while NTK analysis confirms its training dynamics naturally emphasize low-to-mid-frequency signals.
    
    \item We develop a unified $N$-stage pipeline framework that facilitates efficient and flexible hardware implementation of various AFs in INRs. To ensure a fair comparison between different INR architectures, we provide FPGA implementations on the VCU128 platform as well as an ASIC implementation in a 28\,nm process.
    
    \item Experiments across multiple modalities, including images and videos, demonstrate that QuadINR delivers up to 2.06 dB PSNR improvement compared to previous works while simultaneously reducing resource consumption by 65\% and power consumption by 97\%.
\end{itemize}
\section{QuadINR: INR with Quadratic Activation}
Unlike traditional implementations that struggle with sine-based activations, our QuadINR employs a piecewise quadratic activation to reduce complexity while preserving representational power, as shown in Fig.~\ref{fig:function}. It requires only basic multiplication operations instead of complex trigonometric approximations. Our activation is defined as:
\begin{equation}
\phi(x) = 
\begin{cases}
x^2 + 2x, & -2 < x \leq 0 \\
-x^2 + 2x, & 0 < x < 2
\end{cases}
\end{equation}
\label{eqn:activation}
This function is extended periodically with a period of $T=4$ to cover the entire input range, allowing for continuous and efficient signal representation. The smooth transitions between segments ensure continuity and differentiability, which are essential properties for neural network training:
\begin{equation}
\phi'(x) = 
\begin{cases}
2x + 2, & -2 < x \leq 0 \\
-2x + 2, & 0 < x < 2
\end{cases}
\end{equation}
\label{eqn:activation}

The non-linearity and expressive power of the activation function can be analyzed through its Fourier series representation. For the function $\phi(x)$ with period $T=4$, the Fourier series is given by:
\begin{equation}
\phi(x) = a_0 + \sum_{n=1}^{\infty} \left( a_n \cos\left(\frac{2\pi n x}{T}\right) + b_n \sin\left(\frac{2\pi n x}{T}\right) \right)
\end{equation}
where $a_0 = a_n = 0$ since the $\phi(x)$ is an odd function and the $b_n$ is given by
\begin{equation}
\begin{split}
b_n = \frac{2}{T} \int_{-T/2}^{T/2} \phi(x) \sin\left(\frac{2 \pi nx}{T}\right) \, dx= \frac{32}{\pi^3 n^3}
\end{split}
\end{equation}
where $n=2k+1$ and $k \in \mathbb{Z}$. Therefore, the Fourier series of our proposed function can be written as $\phi(x) \approx 1.032 \sin\left(\frac{\pi x}{2}\right) + 0.129 \sin\left(\frac{3\pi x}{2}\right) + \cdots$. The rich harmonic content of this function enables it to capture both low- and high-frequency components, allowing it to approximate a wide range of signals.

We analyzed QuadINR's training dynamics through NTK theory~\cite{ntk2, ntk3}, which tracks how network outputs evolve during training. For a network with output $f_\theta(x) = \phi(\omega_0(Wx + b))$, the NTK is:
\begin{equation}
\Theta(x_1, x_2) = \sum_{\theta} \frac{\partial f_\theta(x_1)}{\partial \theta} \cdot \frac{\partial f_\theta(x_2)}{\partial \theta}
\end{equation}

Comparing QuadINR and SIREN NTKs (where $z_i = W \cdot x_i + b$):
\begin{equation}
\Theta_{Quad}(x_1, x_2) = (2\omega_0^2z_1 + \omega_0)(2\omega_0^2z_2 + \omega_0)[(x_1 \cdot x_2) + 1]
\end{equation}
\begin{equation}
\Theta_{SIREN}(x_1, x_2) = \omega_0^2\cos(\omega_0z_1)\cos(\omega_0z_2)[(x_1 \cdot x_2) + 1]
\end{equation}

The crucial distinction lies in their scaling behavior: QuadINR exhibits polynomial scaling with $\omega_0$ (fourth order), while SIREN shows oscillatory behavior with quadratic $\omega_0$ scaling modulated by cosine terms. This provides QuadINR with more predictable training dynamics while maintaining expressivity.
\section{Hardware Implementation of INR Accelerator}
\subsection{Implementation of INR Accelerator}
\begin{figure}[!t]
\centering
\includegraphics[scale=0.35]{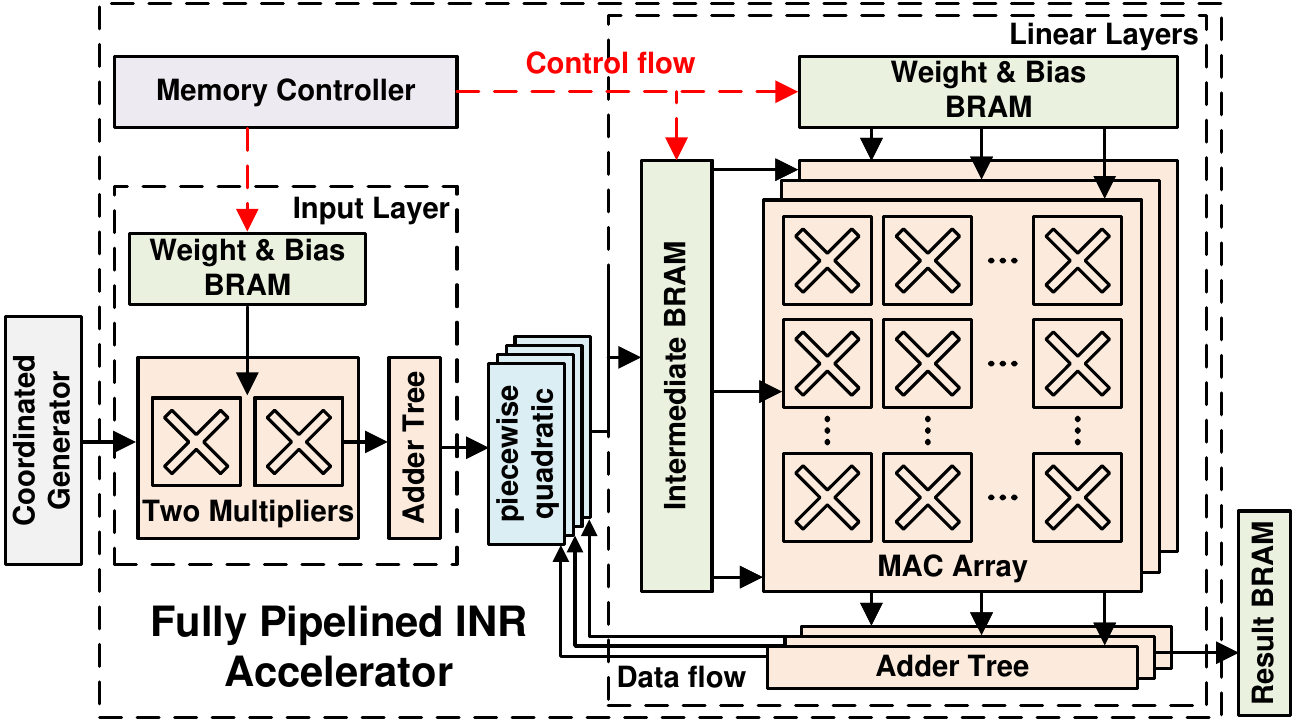}
\caption{Overall architecture of the proposed fully pipelined INR accelerator on FPGA, consisting of a specialized input layer and three linear layers with pipelined processing units for coordinate-to-pixel generation.}
\label{fig_inr_fpga}
\end{figure}

Fig.~\ref{fig_inr_fpga} shows the overall architecture of our proposed Fully Pipelined INR Accelerator, designed to perform group inference in INR applications based on input features. The architecture balances pipelining and parallelism through two main modules corresponding to input dimensionality: the input layer and the linear layer. The input layer maps coordinate inputs to high-dimensional features, while the linear layer generates pixel values through continuous network functions. A dedicated memory control module orchestrates pipeline execution based on internal clock cycles. The linear layer includes 256 parallel MAC arrays, a 256-input addition tree, and two RAM blocks. Weight and bias memory blocks store network parameters, and an intermediate RAM holds partial computation results. Sine activations connect intermediate outputs between layers. 

Structurally similar to the linear layer, the input layer handles only two input activations and uses a smaller MAC array to optimize interlayer pipelining. The output layer omits the activation function and directly stores the generated pixel values in the Result RAM. This parameterized design facilitates network evaluation, while intralayer pipeline computing units enhance the architecture’s efficiency in coordinate-based image generation tasks.
\subsection{Architecture of Different Activation Function Modules}
Hardware resource usage in Taylor expansion implementations~\cite{taylor_1,taylor_2} directly parallels that in CORDIC realizations~\cite{CORDIC1,CORDIC2,CORDIC3}. Since both methods approximate the same activation functions, functions that are resource-intensive in Taylor expansion similarly require more resources with CORDIC, while those with lower demands remain efficient. As the range of X increases, both resource consumption and latency grow to maintain accuracy. Thus, the hardware comparison based on Taylor expansion is applicable to CORDIC, which is why our design uses Taylor expansion to assess the hardware friendliness of various activation functions.

In general, an activation function can be approximated via a Taylor series expansion about its center point (e.g., sinusoidal, Gaussian, etc.). For functions exhibiting variable periodicity of \((|x|+1)x\), the entire expression is first treated as a single variable z and then expanded. For more complex activation functions—such as certain wavelet functions—the Taylor series is applied to each component separately before combining them; the resulting expansions are then multiplied to obtain a composite polynomial. In the case of the sinc function, one must first expand the numerator using a Taylor series and subsequently divide each term by \(x\) to obtain the final polynomial approximation.

Based on this approach, each activation function \( f(x) \) is approximated using its Taylor series expansion up to \( N \) terms, with the coefficients \( c_n \) and the degree \( N \) chosen such that the approximation error remains below 1\% for input values in the range \([-2, 2]\) or \([-1, 1]\). The general form of the Taylor series expansion with even-powered terms or odd-powered terms is expressed as:
\begin{equation}
   f_{\text{even}}(x) \approx \sum_{n=0}^{N} c_{2n}\, x^{2n}, \quad
   f_{\text{odd}}(x) \approx \sum_{n=0}^{N} c_{2n+1}\, x^{2n+1}
\end{equation}
where \( c_n \) represents the coefficients for the \( n \)-th term of the series. Table~\ref{tab:taylor} lists the coefficients and expansion terms for each activation function.

\begin{figure}[!t]
\centering
\includegraphics[scale=0.55]{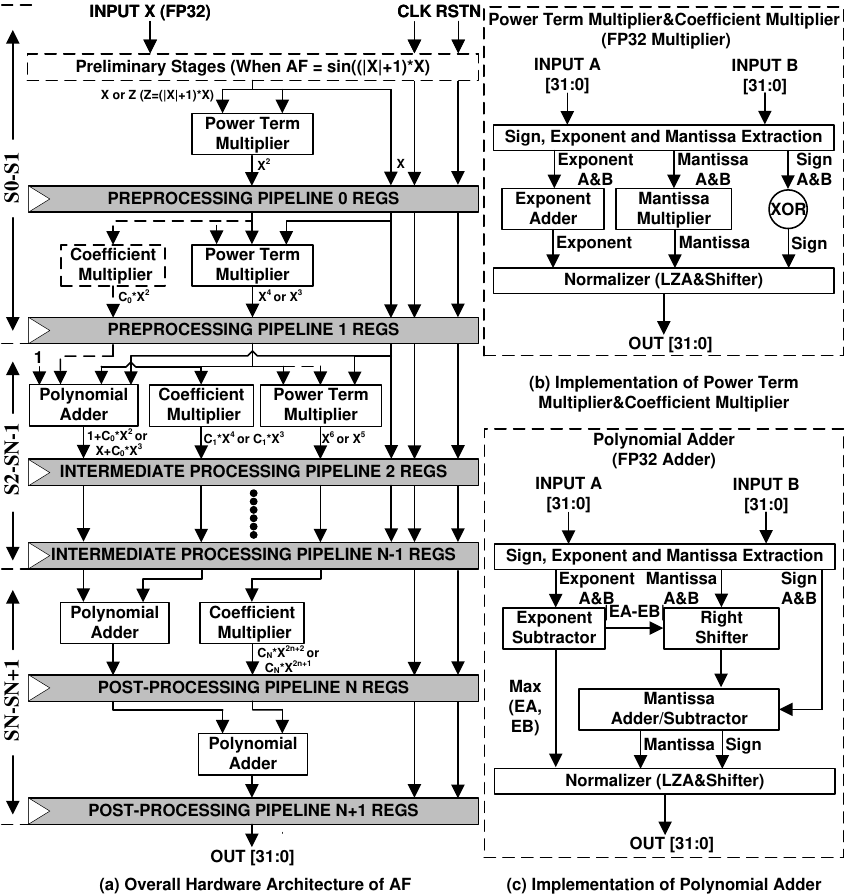}
\caption{Proposed AF Modules: (a) Overall hardware architecture of AF; (b) Implementation of the power term multiplier and coefficient multiplier; (c) Implementation of the polynomial adder.}
\label{fig:af_fpga}
\end{figure}
\begin{table}[!t]
\setlength{\tabcolsep}{1.2pt}
\caption{Comparison of different activation functions using Taylor expansions with an error of less than 1\%.}
\renewcommand{\arraystretch}{1.4}
\centering
\definecolor{gg}{HTML}{e2f0cb}
\begin{tabular}{lcclcc}
\toprule
\multirow{2}{*}[-2.0ex]{\textbf{AF}} & \multicolumn{2}{c}{X $\in$ {[}-2,+2{]}} &  & \multicolumn{2}{c}{X $\in$ {[}-1,+1{]}} \\ \cline{2-3} \cline{5-6} 
 & \textbf{\begin{tabular}[c]{@{}c@{}}Expansion \\ Terms\end{tabular}} & \textbf{\begin{tabular}[c]{@{}c@{}}Max Poly. \\ Degree\end{tabular}} &  & \textbf{\begin{tabular}[c]{@{}c@{}}Expansion \\ Terms\end{tabular}} & \textbf{\begin{tabular}[c]{@{}c@{}}Max Poly. \\ Degree\end{tabular}} \\ \midrule
$\sin(x)$~\cite{siren} & 4 & $x^{7}$ &  & 2 & $x^{3}$ \\
$e^{-x^2}$~\cite{gauss} & 16 & $x^{30}$ &  & 6 & $x^{10}$ \\
$\cos(x)e^{-x^2}$~\cite{wire} & 31 & $x^{60}$ &  & 7 & $x^{12}$ \\
$\sin((|x|+1)x)$~\cite{finer} & 10 & $z^{19}$ &  & 4 & $z^{7}$ \\
$\sin(x)/x$~\cite{sinc} & 4 & $x^{6}$ &  & 3 & $x^{4}$ \\
\rowcolor{gg} $\pm x^2 + 2x$ & 2 & $x^2$ &  & 2 & $x^2$ \\ 
\bottomrule
\end{tabular}
\label{tab:taylor}
\end{table}

\subsection{Circuit Implementation}

The hardware structures of various AFs are depicted in Fig.~\ref{fig:af_fpga}. To ensure consistency in the maximum frequency across different types of activation functions, we employ a novel $N$-stage pipeline architecture for implementing the Taylor-expanded activation functions, as illustrated in Fig.~\ref{fig:af_fpga}(a).

When the activation function (AF) is $\sin((|x| + 1) \cdot x)$, the input $x$ undergoes preprocessing followed by a two-stage pipeline operation. In the first stage, an FP32 adder computes $|x| + 1$, and in the second stage, an FP32 multiplier calculates $(|x| + 1) \cdot x$. The remaining AFs directly enter the standard pipeline processing architecture.

In the $S_0$ stage, the input $x$ enters the Power Term Multiplier to compute the $x^2$ component. As the Taylor expansion exponent increases by $x^2$, this variable continues to pass through the pipeline until the maximum power term is reached. In the S1 stage, the Coefficient Multiplier is represented by a dotted line. For activation functions such as $e^{-x^2}$, $\cos(x) \cdot e^{-|x|^2}$, and $\sin(x)/x$, where the power term of $X$ is even, a Coefficient Multiplier performs coefficient multiplication in this stage. Conversely, for functions like $\sin(x)$ and $\sin((|x| + 1) \cdot x)$, coefficient multiplication is omitted, and the Power Term Multiplier continues to operate in this stage.

In the $S_2$ to $S_{N-1}$ stages, each stage consistently performs coefficient multiplication, power multiplication, and the addition of two terms simultaneously. Upon reaching the SN stage, where the power reaches the maximum expansion level, the penultimate component undergoes coefficient multiplication and the previous term is added. The final $S_{N+1}$ stage performs the last addition.

For the piecewise quadratic AF we propose, only two pipeline stages are necessary. The first stage utilizes both the Power Term Multiplier and the Coefficient Multiplier to compute $x^2$ and $\pm 2x$, while the second stage employs a polynomial unit for direct addition. The hardware implementation demonstrates that the piecewise quadratic AF consumes fewer resources compared to traditional AFs. Detailed implementations of the Power Term Multiplier and Coefficient Multiplier are shown in Fig.~\ref{fig:af_fpga}(b) and Fig.~\ref{fig:af_fpga}(c), respectively. Since our overall network utilizes the FP32 data type, the implementation effectively consists of FP32 multipliers and FP32 adders.

\section{Experiments}
\subsection{Experiment Setup}
Models are trained using PyTorch 1.11.0 on a GeForce RTX 3090 GPU. We train our model using Adam optimizer with a learning rate that decays to 10\% of its initial value over the course of training. We learn representations on high-resolution color images from the Kodak dataset. For video reconstruction, we learn representations on a 5-second video with 25 frames per second. The reconstruction ability is measured by Peak Signal-to-Noise Ratio (PSNR)~\cite{diner}.

We implement various AF modules with SMIC 28nm process library under typical parameters (1.00V, 25$^{\circ}$C, and 1GHz) through the Synopsys Design Compiler. With the netlist file, we extract feature maps from all layers that need to be processed by AFs for use as a testbench. This testbench is then input into Synopsys VCS to generate a waveform diagram, which is subsequently fed into Primetime to accurately characterize the dynamic power consumption of the AF modules. Furthermore, the proposed INR accelerator and various AF modules have undergone functional verification and resource consumption analysis on the VCU128 FPGA at 100 MHz.
\subsection{Experiment Results}
Table~\ref{tab:result} provides a quantitative comparison between our proposed method, QuadINR, and several baseline INR architectures on Kodak image reconstruction tasks. Among the baselines, SIREN achieves PSNR values ranging from 30.52 dB to 36.26 dB, while WIRE shows slightly higher performance, with scores between 30.95 dB and 37.96 dB. FINER outperforms other baselines, attaining the highest PSNR of 40.87 dB on Kodak03. The Gaussian approach delivers results between 29.94 dB and 37.91 dB. In contrast, our proposed QuadINR demonstrates superior performance across the board, achieving 38.86 dB on Kodak03.

\begin{figure}[!t]
\centering
\includegraphics[scale = 0.3]{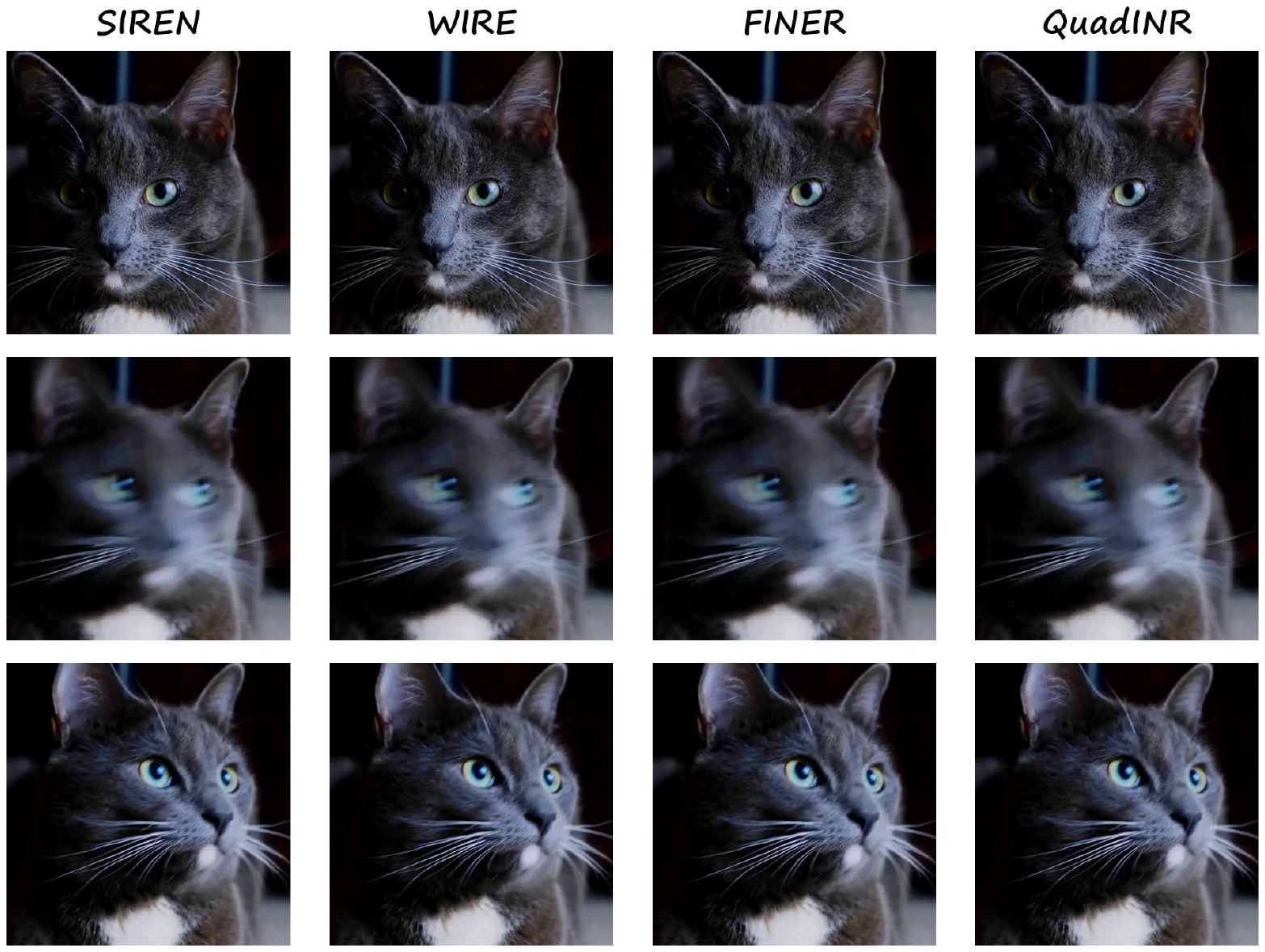}
\caption{Qualitative comparison between SIREN, WIRE, FINER, and our QuadINR in \textbf{video reconstruction}.}
\label{fig:video_result}
\vspace{-0.2cm}
\end{figure}

Fig.~\ref{fig:video_result} demonstrates the visual comparison of different INR architectures on video regression tasks. All methods attempt to reconstruct fine details in the video frames, such as fur texture and facial features. However, there are notable differences in quality among them. While SIREN and WIRE show reasonable reconstruction capabilities, they struggle with preserving the sharpest details. In contrast, our proposed QuadINR method demonstrates superior performance, maintaining better preservation of sharp details and texture consistency. This quality difference is especially evident in challenging areas such as the cat's whiskers and eye regions, where fine-grained detail preservation is crucial for visual fidelity.
\begin{table}[!t]
\centering
\renewcommand{\arraystretch}{1.3}
\caption{Quantitative comparison between INR architectures on Kodak image reconstruction tasks.}
\definecolor{gg}{HTML}{e2f0cb}
\resizebox{\linewidth}{!}{
\begin{tabular}{lcccc}
\toprule
\multirow{2}{*}{\textbf{Methods}} & \multicolumn{4}{c}{\textbf{PSNR}} \\ \cline{2-5}
& \textbf{Kodak01} & \textbf{Kodak02} & \textbf{Kodak03} & \textbf{Kodak04} \\ \hline
SIREN~\cite{siren}                          & 30.52 dB          & 35.21 dB         & 36.26 dB      & 34.52 dB \\
WIRE~\cite{wire}                            & 30.95 dB          & 35.61 dB         & 37.96 dB      & 35.78 dB \\
FINER~\cite{finer}                          & 35.17 dB          & 37.90 dB         & 40.87 dB      & 38.65 dB \\
Gaussian~\cite{gauss}                       & 29.94 dB          & 34.20 dB         & 37.91 dB      & 34.36 dB \\
\rowcolor{gg} QuadINR                       & 32.58 dB          & 36.21 dB         & 38.86 dB      & 36.32 dB \\
\bottomrule
\end{tabular}
}
\label{tab:result}
\vspace{-0.2cm}
\end{table}
\begin{table}[!t]
\renewcommand{\arraystretch}{1.5}
\setlength{\tabcolsep}{1.8pt}
\centering
\definecolor{gg}{HTML}{e2f0cb}
\caption{Performance summary of the proposed INR accelerator and the hardware cost breakdown for the main computation blocks on VCU128 at 100 MHz.}
\label{tab:inr_resources}
\begin{tabular}{lccccccc}
\toprule
\multirow{2}{*}{\textbf{\begin{tabular}[c]{@{}c@{}}Component\\ (Five layers)\end{tabular}}} & \multicolumn{5}{c}{\textbf{Resource Utilization}} & \multirow{2}{*}{\textbf{\begin{tabular}[c]{@{}c@{}}Power \\ (mW)\end{tabular}}} & \multirow{2}{*}{\textbf{\begin{tabular}[c]{@{}c@{}}Latency\\ (ns)\end{tabular}}} \\ \cline{2-6}
 & \textbf{LUT} & \textbf{FF} & \textbf{DSP} & \textbf{LUTRAM} & \textbf{BRAM} &  &  \\ \midrule
MAC Array & 244,759 & 442,268 & 5,130 & 14,394 & 0 & 6,350 & 3,760 \\
Mem. Ctrl. & 60 & 55 & 0 & 0 & 0 & 10 & 80 \\
Storage & 0 & 0 & 0 & 0 & 231 & 615 & 40 \\
AF & 5,032 & 388 & 8 & 0 & 0 & 112 & 80 \\
Others & 525 & 99,015 & 0 & 396 & 0 & 455 & 10280 \\
\rowcolor{gg} Total & 250,376 & 541,726 & 5,138 & 14,790 & 231 & 7,542 & 14240 \\ 
\bottomrule
\end{tabular}
\vspace{-0.2cm}
\end{table}
\begin{table}[!t]
\setlength{\tabcolsep}{1.8pt}
\renewcommand{\arraystretch}{1.4}
\centering
\definecolor{gg}{HTML}{e2f0cb}
\caption{Performance comparison of various AF designs on VCU128 at 100 MHz.}
\resizebox{\columnwidth}{!}{
\begin{tabular}{lccccccccccc}
\toprule
\multirow{3}{*}{\textbf{Design}} & \multicolumn{5}{c}{\textbf{x $\in$ {[}-2,+2{]}}} &  & \multicolumn{5}{c}{\textbf{x $\in$ {[}-1,+1{]}}} \\ \cmidrule{2-6} \cmidrule{8-12} 
 & \multicolumn{3}{c}{\textbf{Resource Utilization}} & \multirow{2}{*}{\textbf{\begin{tabular}[c]{@{}c@{}}Power \\ (mW)\end{tabular}}} & \multirow{2}{*}{\textbf{\begin{tabular}[c]{@{}c@{}}Latency\\ (ns)\end{tabular}}} &  & \multicolumn{3}{c}{\textbf{Resource Utilization}} & \multirow{2}{*}{\textbf{\begin{tabular}[c]{@{}c@{}}Power \\ (mW)\end{tabular}}} & \multirow{2}{*}{\textbf{\begin{tabular}[c]{@{}c@{}}Latency\\ (ns)\end{tabular}}} \\ \cmidrule{2-4} \cmidrule{8-10}
 & \textbf{LUT} & \textbf{FF} & \textbf{DSP} &  &  &  & \textbf{LUT} & \textbf{FF} & \textbf{DSP} &  &  \\ \midrule
SIREN\cite{siren} & 3,868 & 477 & 14 & 112 & 60 &  & 1,312 & 223 & 6 & 35 & 40 \\ 
Gaussian\cite{gauss} & 18,520 & 1,790 & 58 & 520 & 170 &  & 5,835 & 535 & 18 & 153 & 70 \\ 
WIRE~\cite{wire} & 37,069 & 3,682 & 120 & 1,055 & 320 &  & 7,158 & 691 & 24 & 196 & 80 \\ 
FINER~\cite{finer} & 12,150 & 1,334 & 40 & 365 & 140 &  & 4,487 & 572 & 16 & 127 & 80 \\ 
Sinc~\cite{sinc} & 3,433 & 314 & 12 & 85 & 50 &  & 2,206 & 189 & 8 & 49 & 50 \\ 
\rowcolor{gg} QuadINR & 1,258 & 97 & 2 & 28 & 20 &  & 1,258 & 97 & 2 & 28 & 20 \\ 
\bottomrule
\end{tabular}
}
\label{tab:af_resources_fpga}
\vspace{-0.2cm}
\end{table}
\begin{table}[t]
\setlength{\tabcolsep}{1.8pt}
\renewcommand{\arraystretch}{1.4}
\caption{Performance comparison of various AF designs utilizing a 28nm process at 1GHz for 768$\times$512 image processing.}
\centering
\definecolor{gg}{HTML}{e2f0cb}
\resizebox{\columnwidth}{!}{%
\begin{tabular}{lcccclcccc}
\toprule
\multirow{2}{*}[-2.0ex]{\textbf{Design}} &\multicolumn{4}{c}{\textbf{$x \in [-2,+2]$}} &  & \multicolumn{4}{c}{\textbf{$x \in [-1,+1]$}} \\ \cmidrule(lr){2-5} \cmidrule(lr){7-10} 
 & \textbf{\begin{tabular}[c]{@{}c@{}}Area\\ ($\mu$m$^2$)\end{tabular}} & \textbf{\begin{tabular}[c]{@{}c@{}}Static \\ Power (mW)\end{tabular}} & \textbf{\begin{tabular}[c]{@{}c@{}}Dynamic \\ Power (mW)\end{tabular}} & \textbf{\begin{tabular}[c]{@{}c@{}}Energy\\ ($\mu$J)\end{tabular}} &  & \textbf{\begin{tabular}[c]{@{}c@{}}Area\\ ($\mu$m$^2$)\end{tabular}} & \textbf{\begin{tabular}[c]{@{}c@{}}Static \\ Power (mW)\end{tabular}} & \textbf{\begin{tabular}[c]{@{}c@{}}Dynamic \\ Power (mW)\end{tabular}} & \textbf{\begin{tabular}[c]{@{}c@{}}Energy\\ ($\mu$J)\end{tabular}} \\ \midrule
SIREN & 8,415 & 6.83 & 23.30 & 36.79 &  & 3,421 & 3.25 & 8.97 & 14.16 \\ 
Gauss & 37,099 & 20.60 & 112.25 & 177.30 &  & 11,274 & 8.10 & 31.48 & 49.72 \\ 
WIRE & 74,536 & 32.00 & 228.75 & 362.19 &  & 14,287 & 10.10 & 40.45 & 63.89 \\ 
FINER & 25,765 & 15.34 & 69.45 & 109.60 &  & 9,918 & 7.19 & 27.25 & 43.04 \\ 
Sinc & 6,582 & 5.12 & 16.55 & 26.10 &  & 4,092 & 3.23 & 9.50 & 15.00 \\ 
\rowcolor{gg} QuadINR & 1,914 & 1.54 & 6.14 & 9.69 &  & 1,914 & 1.54 & 6.14 & 9.69 \\ 
\bottomrule
\end{tabular}
}
\label{tab:asic_af}
\vspace{-0.2cm}
\end{table}

The hardware cost breakdown of the proposed INR accelerator is
enumerated in Table~\ref{tab:inr_resources}. In the accelerator, not only does the MAC array consume significant resources, but the AF modules—four identical units instantiated for the input and three linear layers—also occupy a considerable portion of the logic. Therefore, optimizing the hardware design of the AF modules is crucial. 
Table~\ref{tab:af_resources_fpga} details the advantages of piecewise quadratic activation functions in hardware implementation on the VCU128 FPGA. When $X \in [-2, 2]$, these functions use only 1258 LUTs, 97 FFs, 2 DSPs, and 28\,mW of energy—resulting in a 65\% to 97\% reduction in resource and power consumption compared to other activation functions such as Sinc and WIRE, while also exhibiting the lowest latency. Moreover, unlike Taylor expansion approximations, the piecewise quadratic activation function achieves zero error. Even within the range $X \in [-1, 1]$, resource and power consumption are reduced by 20\% to 85\% compared with alternatives like SIREN and WIRE.

Table~\ref{tab:asic_af} further demonstrates the hardware advantages under 28\,nm process synthesis, where the piecewise quadratic activation function occupies only 1914\,$\mu$m$^2$, consumes 6.14\,mW of dynamic power, and requires just 9.69\,$\mu$J of energy to process a $768 \times 512$ image—achieving improvements of 60\%–97\% for $X \in [-2, 2]$ and 30\%–85\% for $X \in [-1, 1]$. Together, the FPGA and ASIC data conclusively demonstrate that the piecewise quadratic activation function offers significant hardware benefits, making it the most suitable choice for INR networks in terms of both algorithm performance and hardware implementation.

\section{Conclusion}
We introduce QuadINR, a novel hardware-efficient INR employing a piecewise quadratic activation function that preserves gradient continuity and encompasses diverse frequency components in its Fourier series. A unified $N$-stage pipeline framework is developed to enable efficient hardware implementation of various AFs in INRs, demonstrated with FPGA (VCU128) and ASIC (28nm) implementations. Experiments across multiple modalities show that QuadINR delivers up to 2.06dB PSNR improvement over previous works while simultaneously reducing resource consumption by up to 97\%, power consumption by up to 97\%, and latency by up to 93\%. These results establish QuadINR as an optimal solution for deploying high-quality INRs in resource-constrained environments.

\end{document}